\documentclass{article}
\usepackage{ijcai25}

\usepackage{times}
\usepackage{soul}
\usepackage{url}
\usepackage[hidelinks]{hyperref}
\usepackage[utf8]{inputenc}
\usepackage[small]{caption}
\usepackage{graphicx}
\usepackage{amsmath}
\usepackage{amssymb}
\usepackage{amsthm}
\usepackage{booktabs}
\usepackage{algorithm}
\usepackage{algpseudocode}
\usepackage[switch]{lineno}
\usepackage{tikz}
\usetikzlibrary{fit}
\usepackage[framemethod=TikZ]{mdframed}
\usepackage{stmaryrd}
\usepackage{enumerate} 
\usepackage{colortbl}
\usepackage{bbm}
\usepackage{hyperref}
\usetikzlibrary{calc, positioning, arrows.meta, shapes.geometric, backgrounds}
\usepackage{booktabs} %
\usepackage{natbib}

\usepackage{xcolor} 
\definecolor{paleorange}{RGB}{255, 229, 204} 
\definecolor{palegrey}{RGB}{230, 230, 230}   
\definecolor{paleblue}{RGB}{204, 229, 255}   
\definecolor{palegreen}{RGB}{204, 255, 204}  
\definecolor{palered}{RGB}{255, 204, 204}    

\usepackage[T1]{fontenc}
\usepackage[utf8]{inputenc}
\usepackage{microtype}
\usepackage{inconsolata}
\usepackage{graphicx}

\title{FinMarBa: A Market-Informed Dataset for Financial Sentiment Classification}

\author{
  B. Lefort$^{1,2}$\!\and\!E. Benhamou$^{1,3}$\!\and\!  B. Guez$^1$\!\and\!JJ. Jacques Ohana$^2$\!\and\!E. Setrouk$^2$\!\and\!A. Etienne$^2$\\
  $^1$Ai For Alpha, $^2$Centrale Supélec, $^3$Paris Dauphine PSL, \{first\_name.lasst\_name\}@aiforalpha.com\\
    \vspace{0.3em} \small IJCAI 2025 - FinLLM Workshop - Guangzhou, China, August 28, 2025
}

\begin{document}

\maketitle

\begin{abstract}
Financial sentiment classification typically relies on human-annotated datasets for model fine-tuning and evaluation. However, to the best of our knowledge, existing datasets exhibit inherent biases and do not directly capture market reactions. Despite these limitations, no alternative dataset has been developed specifically for financial sentiment classification based on news.  In this work, we introduce \textbf{FinMarBa}, a dataset that reflects actual market reactions to news, eliminating human-induced biases. We present a fully automated labeling method that enables the construction of large-scale datasets, facilitating both fine-tuning and evaluation of large language models. Since financial sentiment classification is primarily employed for predictive signal generation, we provide empirical evidence demonstrating the quality of our labelling approach compared to conventional human annotation. Furthermore, we release a substantial subset of the dataset on Hugging Face, along with the associated code and fine-tuned models as open-source resources.
\end{abstract}


\section{Introduction}

\textit{"11 August 2010 - VAIAS said today that its net loss widened to EUR4 .8 m in the first half of 2010 from EUR2.3 m in the corresponding period a year earlier."}. This headline extracted from the Financial-Phrasebank dataset \cite{malo2013good} was classified as negative by the human-annotators.  However, on the day following its publication, the stock price of VAIAS increased, signaling a positive market reaction. This discrepancy highlights a fundamental limitation of human-annotated datasets: they often fail to capture the actual impact of financial news on market behavior.

In financial sentiment classification, the primary objective is to construct predictive signals for market movements \citep{7275714, 9175549, malo2013good, lefort2024chatgpt}. To the best of our knowledge, existing datasets for this task are manually annotated by human evaluators \citep{mozetivc2016multilingual, malo2013good}. Typically, a group of individuals—often financial experts or students—is tasked with categorizing financial texts as positive, negative, or neutral based on their subjective interpretation of economic conditions and the expected impact of the news on the stock market.

This approach has two main limitations. First, human annotation introduces biases, as annotators may overlook relevant market conditions when classifying financial news or, conversely, cause data leakage by subconsciously incorporating future information. Second, state-of-the-art language models require extensive training data, yet manually annotating large datasets is both costly and time-consuming. Furthermore, human performance in financial sentiment annotation is subject to variability over time, leading to inconsistencies in classification.

To address these challenges, we propose a methodology for constructing a large-scale dataset that is free from human biases and more accurately reflects market behavior. Recent advancements in natural language processing (NLP) have enabled the use of large language models (LLMs) for extracting and processing financial information \citep{10.1007}. In our work, we leverage GPT-4 to automate the dataset creation process. GPT-4 has demonstrated strong performance in various NLP tasks, including text summarization and entity recognition \citep{mao2023gpteval}.

Our dataset is based on Bloomberg Market Wraps, which offers several advantages over existing datasets that primarily rely on social media sources (e.g., Twitter, Reddit) or region-specific financial texts (e.g., Financial-Phrasebank). Bloomberg provides reliable, high-quality financial news closely followed by market participants. Additionally, its coverage spans global economic events and major financial markets, making it a robust source for financial sentiment classification.

\paragraph{Contributions} 
This work makes several key contributions to financial sentiment classification:

\begin{enumerate}
    \item \textbf{A market-driven annotation framework.} We introduce a structured annotation methodology that classifies financial text based on actual market responses rather than subjective human interpretation.
    
    \item \textbf{The FinMarBa dataset.} We present \textit{FinMarBa}, a large-scale dataset constructed using market-driven annotation. Unlike existing human-labeled datasets, our approach leverages objective financial events, enabling more accurate sentiment classification. The dataset is derived from Bloomberg data, a reliable source that provides comprehensive coverage of global markets and macroeconomic trends.

    \item \textbf{Empirical validation of market-driven annotation.} We provide quantitative evidence demonstrating the advantages of market-based annotation compared to traditional human-labeled approaches. Specifically, we show that the key terms in this dataset align more closely with expected financial market drivers and that the split between positive, negative and indecisive news is more in line with the long term-term positive bias of equity markets as well as the reduction of indecisive news. Additionally, we observe an improvement in backtesting results when using this dataset for financial sentiment analysis with a much higher Sharpe ratio. These findings confirms the effectiveness of market-driven annotation in capturing sentiment signals that are more relevant to financial markets.

    \item \textbf{Facilitating future research.} To promote reproducibility and further advancements in financial NLP, we release a subset of our dataset\footnote{\url{https://huggingface.co/datasets/baptle/financial_headlines_market_based}} and fine-tuned models\footnote{\url{https://huggingface.co/baptle/FinbertMBComparison}} as open-source resources on the Hugging Face platform. We aim to support the broader research community in developing and improving financial sentiment analysis models.
\end{enumerate}

\section{Related Works}

\paragraph{Financial Sentiment Classification} NLP have been widely adopted in the field of finance \citep{araci2019finbert, malo2013good, yang2023investlm}. Sentiment classification of text is one of the most explored task \citep{8334488, kazemian-etal-2016-evaluating}. In addition, prior literature and available dataset for financial sentiment classification are based on a human annotation of the texts \citep{malo2013good}. Sentiment extracted from text are then used for building sentiment signal that are used for building investment strategies \citep{lefort2024stress, electronics12183960}. However, the human annotation does not reflect accurately the likely impact of the news on the market. The human may not be considering the appropriate parameters or may not have enough knowledge (master's student are annotator for the financial\_phrasebank dataset) for classifying correctly the dataset. Also, the existing open-sourced sentiment dataset are not based on fully relevant data, such as tweets or region- or market-focused data. Consequently, a human-biased free dataset where the associated sentiment reflects the real impact on the market is needed.

\paragraph{Financial Sentiment Datasets} 
We identify two types of dataset that are used for financial sentiment classification task. 

\begin{itemize}
    \item \textbf{Financial-Phrasebank dataset.} The Financial-Phrasebank dataset \citep{malo2013good} consists of approximately 5,000 sentences, each annotated with a sentiment label. The dataset primarily focuses on financial news related to northeastern Europe. The annotation process was conducted by a group of 13 individuals, including 10 master's students and 3 researchers, who manually classified the sentences based on their subjective interpretation of economic conditions.
    
    \item \textbf{Social media-based datasets.} Several datasets derived from social media platforms \citep{hussein_2021, Thukral_2023} focus on financial discussions where users express opinions about economic conditions. However, these datasets present two key limitations. First, they comprise a vast amount of content generated by individuals whose opinions may not influence financial markets, as there is no empirical evidence that market participants rely on such sources. Second, sentiment classification in these datasets is often performed using language models trained on semantic features, without considering the underlying economic implications of the expressed sentiments.
\end{itemize}

To address these limitations, we introduce a new dataset that incorporates labels reflecting the real market reactions to financial news, providing a more reliable and objective foundation for financial sentiment classification.

\section{Dataset Construction} \label{sec:dataset_construction}

We collected financial news data from Bloomberg Market Wraps covering the period from 2010 to 2024. These reports, authored by financial experts, provide a concise summary of daily financial events and are widely used in the industry. They serve as a valuable source of information, as they comprehensively cover all major all significant news and events across global markets. 

A key advantage of using Bloomberg Market Wraps is their broad coverage of various markets and regions. This is particularly important for our study, as our objective is to analyze financial sentiment from a global perspective rather than being restricted to a specific geographic area or market segment. These reports are disseminated through multiple channels, including the Bloomberg network as well as online financial platforms such as Yahoo Finance and Investing.com.

In total, we collected over 3,700 daily market wrap reports, each summarizing between 500 and 1,000 individual financial news items, resulting in a corpus exceeding 2 million individual news items. However, instead of analyzing individual news articles, we focus on Bloomberg’s curated daily summaries. This approach leverages expert editorial selection to highlight key financial events and trends, thereby reducing noise and ensuring a dataset that is more representative of market-moving information.

Additionally, we implemented a two-step methodology to further refine the dataset, removing irrelevant content and structuring the data to enhance its usability for financial sentiment analysis.

\subsection{Headline Generation}\label{subsec:headline_generation}
After collecting the daily news, we extracted key headlines that summarize the most significant information of each day. This process enables a more effective condensation of financial reports by emphasizing critical insights while filtering out less relevant details. By structuring the extracted headlines to be concise and non-redundant, we ensure clarity and coherence in the dataset. Additionally, this approach helps reduce noise while preserving essential financial insights, making the extracted headlines particularly relevant for investment decision-making.

The headlines were generated using GPT-4, following a specific prompt designed to extract informative and market-relevant statements. Below is the prompt used in our methodology:

\vspace{0.5cm}

\begin{mdframed}[linecolor=black!20, innerleftmargin=10pt, innerrightmargin=10pt, innertopmargin=10pt, 
innerbottommargin=10pt, backgroundcolor=black!3, roundcorner=3pt]
\textit{You are provided with a financial text, and your task is to extract a list of headlines from it. Each headline must be informative and provide relevant insights for a financial market analyst. Ensure that each headline contains a single piece of information.}
\begin{enumerate}
    \item \textit{Headline for Theme 1}
    \item \textit{Headline for Theme 2}
    \item \textit{...}
\end{enumerate}
\end{mdframed}

This two-step extraction process with first, the filtering of market summaries by Bloomberg’s specialized financial journalists, followed by the extraction of key headlines using a large language model (LLM), enhances  the performance of sentiment classification models by ensuring that the input consists of meaningful, noise-free financial information \cite{lefort2024chatgpt}. At this stage, we obtain a dataset of short, unlabeled financial texts that can be further processed for sentiment analysis.

\subsection{The market-based Annotation Process}\label{subsec:marketbasedprocess}

\paragraph{Ticker Identification}
Firstly, we use GPT-4 to identify and assign a list of tickers associated with each headline. The model has demonstrated efficiency in accurately determining the involved entities in a text \cite{wu2023bloomberggpt}. By doing so, we identify the markets which are the most likely to react to the news. Indeed, a headline may concern specific markets and is unlikely to provoke a reaction on all the markets at once. This step ensures that the associated sentiment reflects the most precisely the true market conditions. In this setup, it's necessary to note that the headline is considered to have an equal impact on all the related tickers. This does not affect the labelling of the data since we want to capture the macro effect of the headline. If the majority of the associated markets have the same common sentiment, then the macro sentiment will be correctly evaluated. \cite{lefort2024chatgpt} detailed in their work that it exists a macro impact of the news on the markets. It remains that even if it seems accurate after double-checking by financial experts, we are aware that the hypothesis whereby headlines are entirely responsible for market movements cannot be completely dismissed. Table~\ref{tab:ticker_distribution} provides a subset of our broader sixty thousands headlines, illustrating specific data entries and their associated financial tickers. The table aligns dates with corresponding news headlines and the related financial market tickers, providing a snapshot from January 4, 2010, to January 31, 2024. Entries include significant financial events, such as shifts in the dollar value, stock market highs, oil price changes, etc.

\begin{table}[!htbp]
    \centering
    \caption{Sample of our large Dataset and corresponding list of Ticker(s)}
    \label{tab:ticker_distribution}
    \resizebox{\columnwidth}{!}{
    \begin{tabular}{llc}
        \toprule
        \bfseries Date & \bfseries Headline & \bfseries Tickers \\
        \midrule
        2010-01-04 & Dollar Slumps Amid Worldwide Manufacturing Improvement & [UUP, ACWI] \\
        \ldots & \ldots & \ldots \\
        2022-08-17 & S\&P 500 Rises 1\% to All-Time High, Treasuries Lose Gains & [GSPC, TNX] \\
        \ldots & \ldots & \ldots \\
        2024-01-31 &  West Texas Intermediate Crude Falls 1.3\% to \$76.83 a Barrel & [CL=F]\\
        \bottomrule
    \end{tabular}
    }
\end{table}

Table~\ref{tab:ticker_distribution} shows the distribution of key tickers across the dataset. Notably, the most frequent tickers are from major equity markets, particularly the US, highlighting its dominance in global finance and extensive media coverage.

\begin{figure}[!htbp]
    \centering
    {{\includegraphics[scale=0.45]{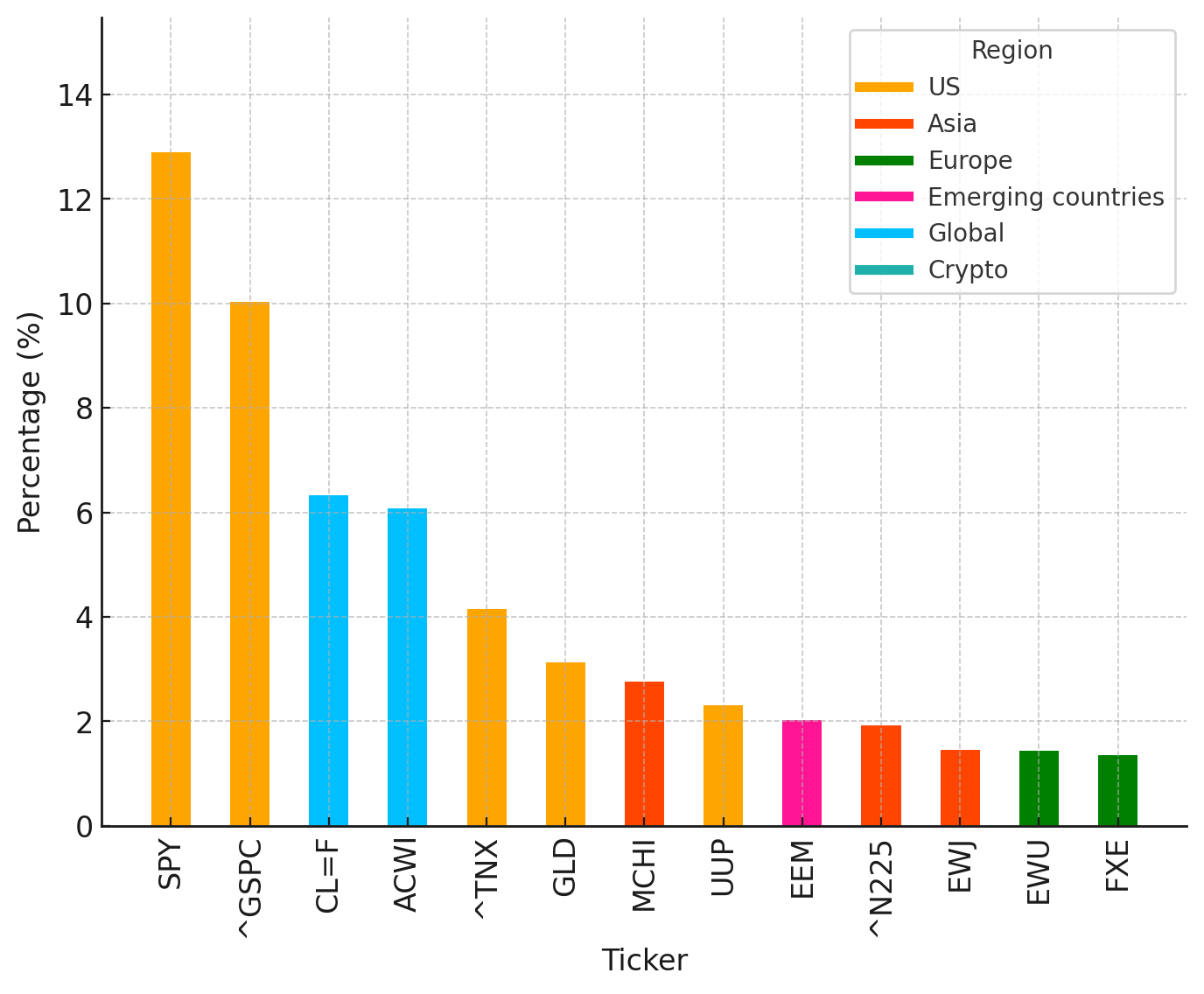} }}%
    \caption{Distribution of the main tickers by region. The sum of these tickers' percentage is not 100\% as only the most represented are considered.}\label{fig:matrix_grid}%
\end{figure}

The table predominantly features equity-related tickers, which corresponds with the sentiment analysis's goal to evaluate feelings toward significant equity sectors. The main region is the US, which seems very consistent since the American market is the largest and most influential. Moreover, the analysis intriguingly links the global market sentiment not only with traditional equities but also with diverse asset classes including Bitcoin, bonds, and gold, showcasing a multifaceted approach to understanding market dynamics. Even if we only showcase in table~\ref{tab:ticker_distribution} less than half of the tickers, the table underscores the complexity and interconnectedness of modern financial markets, with over five thousands tickers and multiple underlying types (equities, bonds, commodities, credit, currencies and even crypto-currencies).

\paragraph{Ticker's Related Market Reaction}
The algorithm classifies financial headlines based on their impact on stock prices using a quantile approach. Historical stock performance data is utilized to categorize the impact of news on the stock prices in a systematic manner.

For each stock ticker, denoted as \( T_k \), the algorithm operates in the following manner:

\begin{enumerate}
    \item \textbf{Historical Data Retrieval:} Acquire historical closing prices for the past five years, maintaining a rolling window approach—considering the history up to the publication date of the relevant financial headline.
    \item \textbf{Percentage Change Calculation:} Compute the daily percentage change in closing prices the day after the publication of the headline, denoted as \(\Delta P_{T_k}\).
    \item \textbf{Quantile Determination:} Determine the specific quantiles \( Q_{0.3, T_k} \) and \( Q_{0.6, T_k} \), representing the 30\% and 60\% thresholds, respectively, based on the rolling historical price data.
    \item \textbf{Classification:} Assign a category to the impact of the news—positive (\(+1\)), negative (\(-1\)), or neutral (\(0\))—by comparing the calculated \(\Delta P_{T_k}\) against the quantiles.
\end{enumerate}

The classification function, \( C(T_k, \Delta P_{T_k}) \), is defined by the following:

\begin{equation}
C(T_k, \Delta P_{T_k}) = 
\begin{cases} 
+1 & \text{if } \Delta P_{T_k} > Q_{0.6, T_k}, \\
-1 & \text{if } \Delta P_{T_k} < Q_{0.3, T_k}, \\
0 & \text{otherwise}.
\end{cases}
\end{equation}

\begin{figure}[!htbp]
\begin{center}
\begin{tikzpicture}[>=Stealth, scale=0.8] 
    \draw[-, line width=1pt, dotted] (-5,0) -- (-4.5,0); 
    \draw[->, line width=1pt] (-4.5,0) -- (4.5,0); 
    
    \draw[-, line width=1pt] (-4.5,0.5) -- (-4.5,-0.5) node[below] {$t-1250$};
    \draw[-, line width=1pt] (2.5,0.5) -- (2.5,-0.5) node[below] {$t$};
    \draw[-, line width=1pt] (3.2,0.5) -- (3.2,-0.5) node[below] {$t+1$};
    
    \fill[opacity=0.3, blue!50] (-4.5,0) rectangle (2.5,0.3);
    \fill[opacity=0.3, paleblue] (2.5,0) rectangle (3.2,0.3);
\end{tikzpicture}
\end{center}
\caption{Time ordering of the automatic labelling steps. The headline is published at time $t$, the historical price distribution is gathered from $t-5$ years to $t$ and the next day price return is from $t$ to $t+1$. }
\label{fig:time_step_process}
\end{figure}
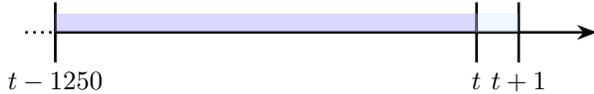

The sequence of steps over time plays a crucial role, as depicted in Figure~\ref{fig:time_step_process}. For a given headline published at time $t$, we examine the preceding five-year price history of the associated ticker, delineated in navy blue, up to the publication date. This historical percentage change is used to compare the subsequent day's return (from $t$ to $t+1$) against the historical distribution. This five-year span is chosen to align with our intent to evaluate the immediate effects of news while capturing a substantial portion of the ticker $T_k$'s past return distributions.

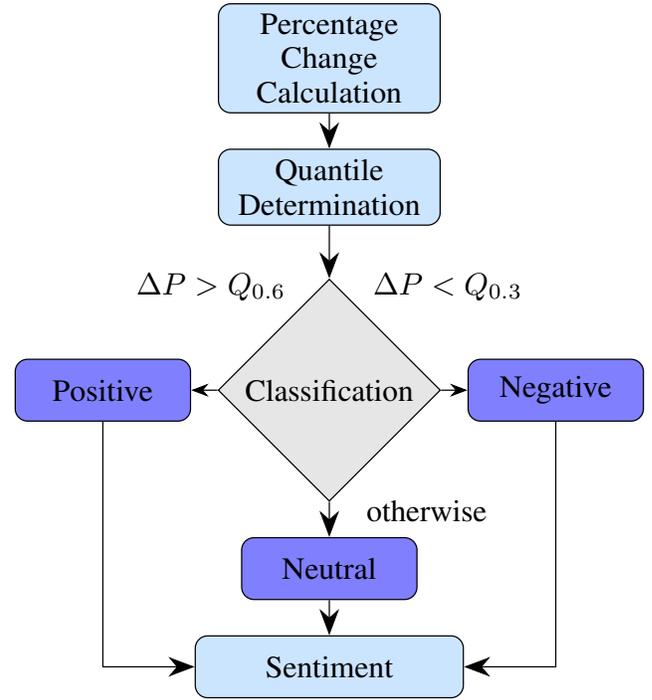
\begin{figure}[!htbp]
\resizebox{\columnwidth}{!}{
\centering
\begin{tikzpicture}[>=Stealth, 
    node distance=0.4cm and 0.3cm, 
    block/.style={rectangle, draw, fill=paleblue, text width=6.5em, text centered, rounded corners, minimum height=2em},
    label/.style={rectangle, draw, fill=blue!50, text width=5.0em, text centered, rounded corners, minimum height=2em},
    decision/.style={diamond, draw, fill=palegrey, text width=6.5em, text badly centered, node distance=0.6cm, inner sep=0pt},
    line/.style={draw, -{Stealth[length=3mm]}},
    line2/.style={draw, -{Stealth[length=2mm]}},
    auto]

    \node[block] (calculation) {Percentage Change Calculation};
    \node[block, below=of calculation] (quantile) {Quantile Determination};
    \node[decision, below=of quantile] (classification) {Classification};
    \node[label, left=of classification] (positive) {Positive};
    \node[label, right=of classification] (negative) {Negative};
    \node[label, below=of classification] (neutral) {Neutral};
    \node[block, below=of neutral, text width=8em] (global) {Sentiment};
    
    \path [line] (calculation) -- (quantile);
    \path [line] (quantile) -- (classification);
    \path [line2] (classification) -- node [near start, above=0.9cm] {$\Delta P > Q_{0.6}$} (positive);
    \path [line2] (classification) -- node [near start, above=0.9cm] {$\Delta P < Q_{0.3}$} (negative);
    \path [line] (classification) -- node [near start, right=0.3cm] {otherwise} (neutral);
    \path [line] (positive.south) |- (global);
    \path [line] (negative.south) |- (global);
    \path [line] (neutral) -- (global);
    
\end{tikzpicture}
}
\caption{Automatic Classification of Financial Headlines. Blue blocks represent data processing steps, green block represents decision points for classification and orange blocks corresponding labels.
}\label{fig:label_process}
\end{figure}

This framework provides insight into news impact, using historical volatility and performance benchmarks. The figure~\ref{fig:label_process} summarizes the process and explain that we start from an initial calculation of the percentage change of the identified ticker used to determine a quantile and then convert this quantile into a sentiment score that is either 'Positive' if the percentage change exceeds the 60th percentile, 'Negative' if below the 30th percentile, and 'Indecisive' otherwise. \\

\begin{figure*}[!htbp]
\centering
\resizebox{0.9\textwidth}{!}{ 
\begin{tikzpicture}[>=Stealth, node distance=1cm and 0.8cm,
    block/.style={rectangle, draw, fill=paleblue, text width=8em, text centered, rounded corners, minimum height=4em},
    newblock/.style={rectangle, draw, fill=blue!50, text width=8em, text centered, rounded corners, minimum height=4em},
    imageblock/.style={rectangle, draw, text width=8em, text centered, minimum height=2.5cm}, 
    line/.style={draw, -{Stealth[length=2mm]}},
    group/.style={rectangle, draw=blue, dashed, thick, inner sep=0.2cm, rounded corners},
    auto]

    \node[block] (collect) {Collect News Headlines};
    \node[block, above right=0.5cm and 1cm of collect] (human_read) {Human \\Read};
    \node[block, right=1.5cm of human_read] (human_interp) {Human \\Interpret};
    \node[block, right=1.5cm of human_interp] (human_label) {Human \\Label};

    \node[newblock, below right=0.5cm and 1cm of collect] (ticker_id) {Identify Ticker};
    \node[newblock, right=1.5cm of ticker_id] (ticker_compute) {Evaluate Ticker Return};
    \node[newblock, right=1.5cm of ticker_compute] (ticker_label) {Machine \\Label};

    \node[block, below right=0.5cm and 1cm of human_label] (evaluate) {Financial Sentiment \\Dataset}; 

    \node[group, fit=(human_read) (human_interp) (human_label)] {};

    \node[imageblock, below=1cm of ticker_label] (image_block_machine) {\makebox[8em]{\includegraphics[scale=0.20]{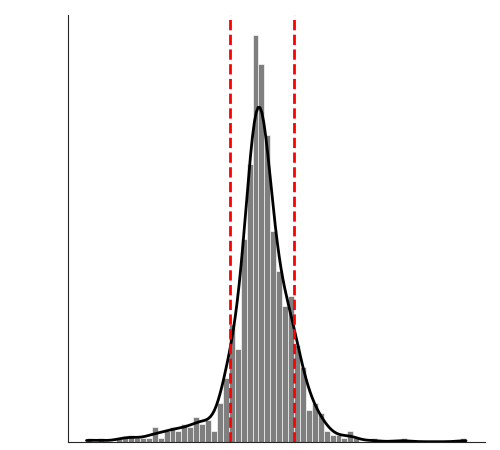}}};

    \node[imageblock, below=1cm of ticker_compute] (image_block_evaluate) {\makebox[8em]{\includegraphics[scale=0.2]{images/dist_return.png}}};

    \path [line] (collect) |- (human_read);
    \path [line] (human_read) -- (human_interp);
    \path [line] (human_interp) -- (human_label);

    \path [line] (collect) |- (ticker_id);
    \path [line] (ticker_id) -- (ticker_compute);
    \path [line] (ticker_compute) -- (ticker_label);

    \path [line] (human_label) -| (evaluate);
    \path [line] (ticker_label) -| (evaluate);

    \path [line] (ticker_label) -- (image_block_machine);
    \path [line] (ticker_compute) -- (image_block_evaluate);

    \node[above right=1.8cm and 0.8cm of collect, align=center] (current_approach) {Current\\Approach};
    \node[below right=1.8cm and 0.8cm of collect, align=center] (new_approach) {New\\Approach};

\end{tikzpicture}
}
\caption{Full Process of Dataset Annotation.}
\label{fig:full_process_sum}
\end{figure*}

Our approach of building an automatic annotator for sentiment classification is new to the financial sector and has not been explored in the academic literature. Unlike other industries, the initial step involves accurately labeling each news item, which is not a simple task. To summarize our contribution, we provide a recap of the entire process, including the new steps, as shown in figure~\ref{fig:full_process_sum}. The dotted light-blue box denotes the usual steps of annotating financial news in the sentiment classification task. They all presents human biases and are slow to achieve. The given labels might not be accurate and reflect the real global market reaction because of the human's sensitivity when reading. The dark blue boxes are the new steps that proposes an alternative to avoid human biases and automate the annotation process.

\section{The FinMarBa Dataset}

After following the described process in section~\ref{sec:dataset_construction}, we obtained a dataset with a total of 61,252 annotated headlines. These headlines have been extracted from a daily news from 01-01-2010 to 31-01-2024. In this section, we focus on FinMarBa's improvements over the dataset most commonly used for this task to date (the Financial-PhraseBank dataset \footnote{\url{https://huggingface.co/datasets/takala/financial_phrasebank}}). We start by describing the improvements from a statistical point of view.

\paragraph{The Annotation Consistency}
The distribution of the sentiment is expected to be quite balanced between the three classes: positive, negative and indecisive. If one sentiment is overwhelmingly dominant, we expect the market to follow that trend and not reflect any other sentiment. However, given that all major equity markets have a slight long-term positive bias, we also expect positive headlines to be slightly more represented. Table~\ref{tab:sentiment_comparison} is consistent with the expected behavior for both on the data set, producing a fairly balanced classification. The main difference is that the FinMarBa dataset reflects this positive market bias, while the Financial-Phrasebank dataset does not.

\begin{table}[!htbp]
  \centering
  \resizebox{\columnwidth}{!}{ 
  \begin{tabular}{lcc}
    \hline
    \textbf{Sentiment} & \textbf{Financial-Phrasebank (\%)} & \textbf{FinMarBar (\%)} \\
    \hline
    Positive & 28.13 & 42.11 \\
    Negative & 12.46 & 31.43 \\
    Indecisive & 59.41 & 26.45 \\
    \hline
  \end{tabular}
  }
  \caption{Proportion of sentiment labels in Financial Phrasebank and Market-based datasets.}
  \label{tab:sentiment_comparison}
\end{table}

Thus, the distribution of labels in the FinMarBa dataset is more consistent with actual market behavior.

\begin{figure}[!htbp]
    \centering
   \resizebox{0.75 \columnwidth}{!}{ 
    {{\includegraphics[scale=0.3]{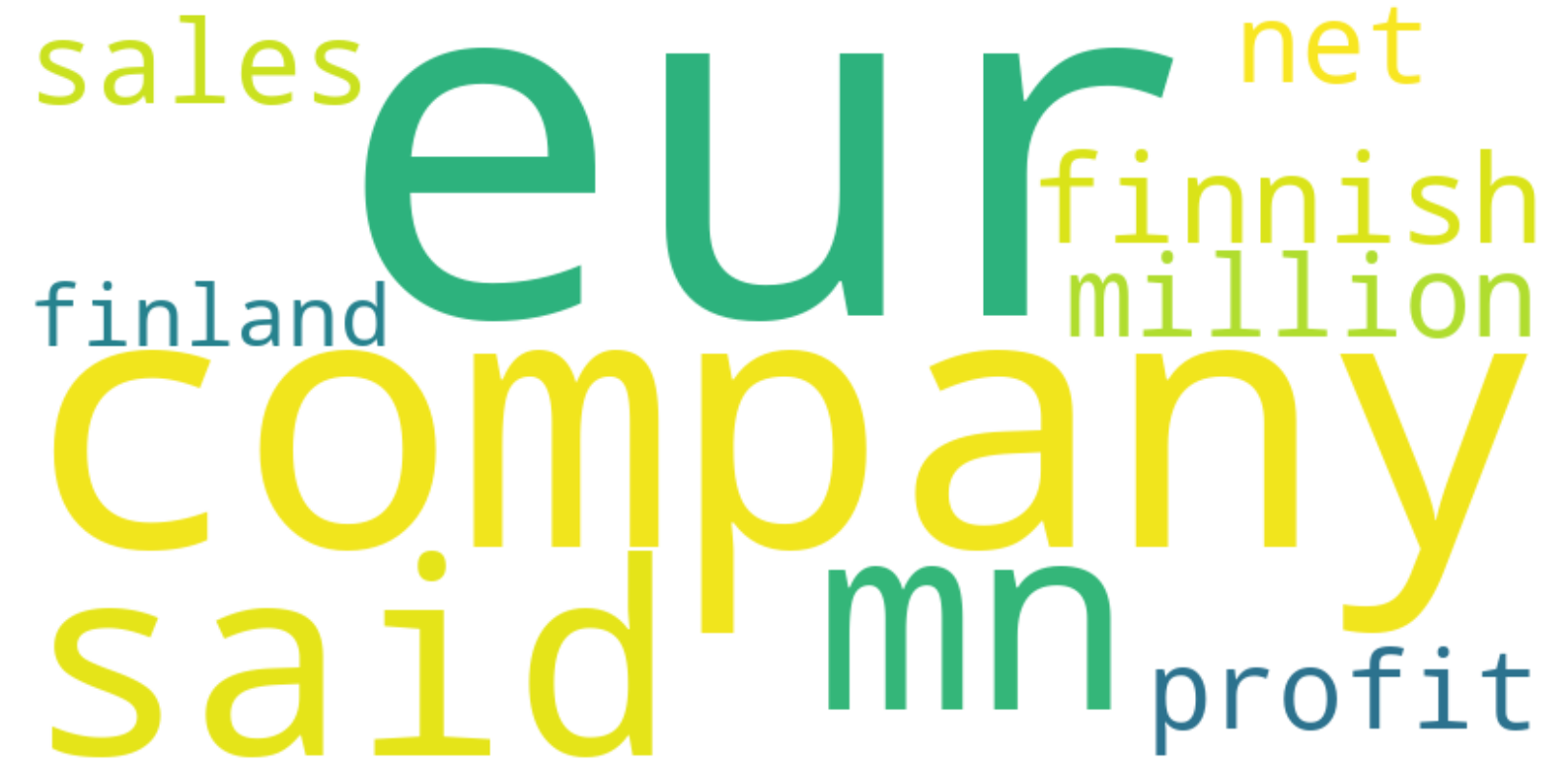} }}%
    }
    \caption{Top 10 words in Financial-Phrasebank}\label{fig:word_cloud_fb}%
\end{figure}

\begin{figure}[!htbp]
    \centering
   \resizebox{0.75 \columnwidth}{!}{ 
    {{\includegraphics[scale=0.5]{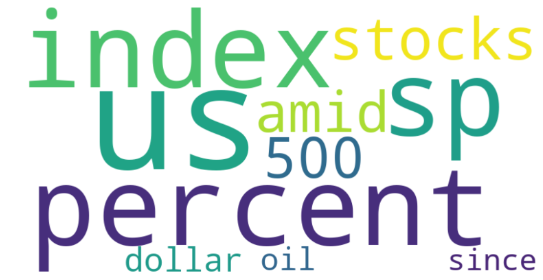} }}%
    }
    \caption{Top 10 words in FinMarBa }\label{fig:word_cloud_mb}%
\end{figure}

\paragraph{Regional Coverage Comparison} 
To construct a predictive macro signal for equity markets, headlines should predominantly cover the U.S., the largest and most influential market, while maintaining balanced coverage of other regions.

Figure~\ref{fig:word_cloud_fb} shows the most frequent words in the Financial-Phrasebank dataset, revealing a strong eurozone focus, with high occurrences of "EUR," "Finland," and "Finnish," indicating a regional bias toward northeastern Europe. 

In contrast, Figure~\ref{fig:word_cloud_mb} highlights the most frequent terms in the FinMarBa dataset, where U.S.-related words dominate, aligning with expectations. The prominence of "stocks," "dollar," and "oil" further reflects key economic drivers shaping global markets.

The FinMarBa dataset better reflects the historical positive bias of the equity market and provides better global coverage than the Financial-Phrasebank dataset. The next section focuses on providing quantitative evidence of the quality of the FinMarBa annotation compared to the Financial-Phrasebank annotation.

\section{Proof of the FinMarBa Annotation Contribution}

When it comes to financial sentiment classification, the main objective is to create a signal derived from the model's classification. The task of classifying sentiments should not be understood on the basis of semantics, but on the basis of market perception. The signal is expected to be predictive of the markets. The intuition is that if the model predicts without error on all the labels, then the signal generated will perform very well when used in the markets. 
To assess the quality of the signal generated from the FinMarBa annotation, we compared it with the Financial-Phrasebank dataset. The methodology employed involved two steps:
\begin{enumerate}
    \item \textbf{Model Fine-Tuning} We train a BERT model on the FinMarBa dataset.
    \item \textbf{Backtest on the S\&P500} We backtest the S\&P500 from 2019 to 2024 to assess the signal generated by the models.
    \item \textbf{Robustness} To validate the quality of the classification, we carried out several experiments focusing on signal robustness.
\end{enumerate}

The Financial-Phrasebank was used to train the well-known FinBERT model \citep{araci2019finbert}. This model will serve as a basis for comparison between the two datasets.

\subsection{Model Fine-Tuning}
First, we divided the data from 2010 to 2019 into a training set and the data from 2019 to 2024 into a test set (representing 4.5 years). We chose 2019 because we wanted to ensure that no training data for FinBERT would be used in the future for FinMarBaBERT. We selected exactly the same hyperparameters as those detailed in the paper published on FinBERT. The result is two models based on BERT: FinBERT is trained on the Financial-Phrasebank dataset and FinMarBaBERT is trained on the FinMarBa dataset. Both models have been trained under the same conditions and with the same hyperparameters. The only difference lies in the dataset used for training. 

\subsection{Backtest Results on the S\&P500}

\paragraph{The Sharpe Ratio} 
To evaluate financial strategies, we use the Sharpe Ratio (\(SR \)), a standard metric for risk-adjusted returns defined as:

\begin{equation}
SR = \frac{\mathbb{E}[R]}{\sigma}
\end{equation}

where \( \mathbb{E}[R] \) is the expected return of the strategy, while \( \sigma \) is the standard deviation of the strategy's excess return, representing risk. A higher Sharpe Ratio indicates superior risk-adjusted performance, making it a key metric for assessing financial models.

\paragraph{The Signal} From the sentiment on each headline, we derive a daily score that is detailed precisely in \citep{lefort2024chatgpt}. It is computed as follows.

The sentiment score \( S \) for a day with \( N \) headlines is given by:
\begin{equation}
S = \frac{\sum_{i=1}^{N} p(h_i) - \sum_{i=1}^{N} n(h_i)}{\sum_{i=1}^{N} p(h_i) + \sum_{i=1}^{N} n(h_i)} 
\end{equation}

Finally, we obtain two sets of data. One for FinMarBaBERT and the other for FinBERT.

\begin{table}[!htbp]
  \centering
  \begin{tabular}{lccc}
        \hline
        \textbf{Date} & \textbf{Positive} & \textbf{Negative} & \textbf{Score} \\
        \hline
        2019-12-12 & 11 & 3 & 0.57\\
        2019-12-13 & 6 & 6 & 0.00 \\
        \multicolumn{4}{c}{\centering\ldots} \\
        2023-12-01 & 8 & 3 & 0.45 \\
        \hline
    \end{tabular}
  \caption{Dataset example containing the daily score.}
  \label{tab:dataset_score}
\end{table}

\paragraph{Results} 
Table~\ref{tab:result_backtest} shows that FinMarBa outperforms Financial-Phrasebank, with a Sharpe Ratio of 0.30 vs. -0.13. This highlights the stronger predictive power of market-driven annotation.T-statistics, obtained by multiplying Sharpe ratios by \( \sqrt{4.5 \times 250} \), confirm statistical significance with p-values below 1 percent. The negative Sharpe ratio suggests that the signal is driven by noise rather than meaningful sentiment insights.

\begin{table}[!htbp]
  \centering
  \resizebox{\columnwidth}{!}{
  \begin{tabular}{lcc}
    \hline
    \textbf{Dataset} & \textbf{Sharpe Ratio} &  \textbf{T-stat (p-value) }\\
    \hline
    FinMarBa & 0.30 & 10 (0) \\
    Financial-Phrasebank & -0.13 & -4.36 (1.3 e-5) \\
    \hline
  \end{tabular}
  }
  \caption{Sharpe Ratios of sentiment-based signals from FinMarBa and Financial-Phrasebank. Higher values indicate stronger risk-adjusted returns, with FinMarBa outperforming Financial-Phrasebank.}
  \label{tab:result_backtest}
\end{table}

\subsection{Signal Robustness}
To validate further the consistency and reliability of our market-based labeling method, we conducted a series of robustness tests. These experiments were designed to assess how the signal's performance changes when we introduce controlled perturbations to the headline data.

\paragraph{Methodology} 
We applied a sliding time window approach, where daily headlines were randomly exchanged within a predefined window. This process was repeated for different window sizes and exchange rates, considering both past (backward-looking) and future (forward-looking) time frames. If our labeling method accurately captures market sentiment, forward-looking windows should yield higher Sharpe ratios than backward-looking ones, particularly at higher exchange rates. Increasing exchange rates in future windows should further improve performance by incorporating more forward-looking information, while past-looking windows should degrade performance as noise increases.

\paragraph{Experimental Setup} 
We tested window sizes of 5, 10 and 15 days, with exchange rates ranging from 10\% to 50\%. Experiments were conducted in forward-looking (future) direction. Then we computed the difference in Sharpe ratio between the FinMarBa and Financial-Phrasebank dataset. A positive difference would indicate that the FinMarBa dataset has better labelling than the Financial-Phrasebank.

\paragraph{Results Consistency} 
The findings of our robustness tests are summarized in Table~\ref{tab:sharpe_difference}, which reports the differences in Sharpe ratios between the FinMarBa dataset and the Financial-Phrasebank dataset across various window sizes and exchange rate thresholds. The results indicate that FinMarBa consistently achieves higher Sharpe ratios, suggesting improved predictive performance. Sharpe ratios calculated over 4.5 years are statistically significant at a 5 percent confidence level, corresponding to a critical value of 1.65 from the inverse normal distribution. A ratio is significant if it exceeds \( \frac{1.65}{\sqrt{4.5 \times 250}} \approx 0.05 \), confirming that all reported values are statistically significant.
The performance gap widens as the threshold increases, with differences rising from 0.50 at 10\% to 1.94 at 50\% for a 5-day window. A similar trend is observed for 10-day and 15-day windows, confirming that greater access to future labels enhances predictive performance. 
The sharpest increase occurs in shorter windows, supporting the hypothesis that improved foresight amplifies dataset differentiation.

\begin{table}[!htbp]
    \centering
    \caption{Difference of Sharpe Ratios between FinMarBa and Financial-Phrasebank}
    \label{tab:sharpe_difference}
    \resizebox{\columnwidth}{!}{
    \begin{tabular}{lccccc}
        \toprule
        \textbf{Window Size} & \textbf{10\% Rate} & \textbf{20\% Rate} & \textbf{30\% Rate} & \textbf{40\% Rate} & \textbf{50\% Rate} \\
        \midrule
        5 days  & 0.50 & 1.03 & 0.77 & 1.52 & 1.94 \\
        10 days & 0.45 & 0.69 & 0.61 & 1.22 & 0.62 \\
        15 days & 0.52 & 0.77 & 0.44 & 0.58 & 0.39 \\
        \bottomrule
    \end{tabular}
    }
\end{table}

\section{Conclusion}

In this paper, we introduced FinMarBa, a novel market-informed dataset for financial sentiment classification. Our approach addresses the limitations of existing human-annotated datasets by directly reflecting market reactions to news, thereby eliminating human bias and providing a more accurate representation of financial sentiment.

The key contributions of our work include:

\begin{enumerate}
    \item A detailed protocol for market-based annotation of financial texts, leveraging advanced NLP techniques and real market data.
    \item The creation of the FinMarBa dataset, which offers a comprehensive and unbiased view of financial sentiment across various markets and regions.
    \item Quantitative evidence demonstrating the superiority of our market-based annotation method over previous traditional human-annotated datasets.
    \item Open-source release of a significant sample of our dataset and fine-tuned models, facilitating further research and improvements in the field.
\end{enumerate}

Our experimental results show that FinMarBa consistently outperforms the widely-used Financial-Phrasebank dataset in terms of sentiment distribution, global market coverage, and predictive power. The robustness tests further validate the consistency and reliability of our labeling method, particularly in capturing short-term market sentiments and maintaining performance under various perturbations.

The implications of this work are significant for both academic research and practical applications in finance. FinMarBa provides a more accurate foundation for developing and evaluating financial sentiment classification models, which can lead to improved predictive signals for market analysis and investment strategies.

Future work could explore the application of FinMarBa to other financial markets and asset classes, as well as investigating the integration of this dataset with more advanced machine learning techniques. Additionally, continuous updating of the dataset with new market data could ensure its ongoing relevance and effectiveness.

In conclusion, FinMarBa represents a significant advancement in financial sentiment analysis, offering a more reliable and market-aligned alternative to existing datasets. We believe this contribution could foster new developments in financial NLP and support the advancement of more accurate and reliable sentiment analysis methodologies.

\section*{Limitations}

While FinMarBa offers a novel approach to financial sentiment classification by utilizing market reactions for dataset labeling, several limitations should be acknowledged. Firstly, the reliance on market data means that the dataset is primarily reflective of publicly traded companies, potentially excluding sentiments related to private entities or sectors with less market transparency. Secondly, the automated labeling process, though designed to reduce human bias, may not fully capture the nuanced contexts of certain financial news, leading to potential misclassifications. Additionally, the dataset's focus on historical market reactions may not account for unprecedented events or shifts in market dynamics, which could affect the generalizability of models trained on this data. Finally, while efforts have been made to ensure data quality, the dependence on sources like Bloomberg Market Wraps means that any inherent biases or inaccuracies in these sources could propagate into the FinMarBa dataset.

\clearpage

\bibliographystyle{named}
\bibliography{main}

\end{document}